\documentclass[10pt,twocolumn,letterpaper]{article}
\usepackage[pagenumbers]{cvpr}
\usepackage{array}
\usepackage{tabularx}
\usepackage{placeins}
\usepackage{needspace}

\newcommand{\pmstd}[1]{\,{\scriptsize$\pm$#1}}
\newcommand{\ablrow}{\hspace{0.3em}\rotatebox[origin=c]{180}{$\Lsh$}\,}
\newlength{\paraleadskip}
\makeatletter
\newcommand{\paralead}[1]{\par\if@nobreak\else\Needspace{4\baselineskip}\vspace{\paraleadskip}\fi
  \noindent\textbf{#1}\enspace\ignorespaces}
\newcommand{\paragap}{\par\if@nobreak\else\vspace{\paraleadskip}\fi\noindent\ignorespaces}
\makeatother

\usepackage[most]{tcolorbox}
\usepackage{fvextra}
\definecolor{promptblue}{HTML}{0061C3}
\definecolor{promptblueback}{HTML}{F3F7FE}
\definecolor{promptgreen}{HTML}{10855A}
\definecolor{promptgreenback}{HTML}{F0F8F4}
\definecolor{promptgray}{HTML}{6B7280}
\definecolor{promptgrayback}{HTML}{F7F7F8}
\newtcolorbox{promptbox}[2][]{enhanced, breakable, boxrule=0.5pt, arc=1.5pt,
  left=4pt, right=4pt, top=3pt, bottom=4pt, toptitle=1.5pt, bottomtitle=1.5pt,
  fonttitle=\sffamily\bfseries\footnotesize, coltitle=white,
  fontupper=\footnotesize\frenchspacing\setlength{\parskip}{1.5pt}\raggedright,
  before skip=6pt, after skip=8pt, title={#2}, #1}
\newcommand{\prompthead}[1]{\par\vspace{3pt}\noindent{\sffamily\bfseries\color{black!75}#1}%
  \par\nopagebreak}
\newcommand{\promptitem}{\par\hangindent=0.8em\hangafter=1\noindent}
\DefineVerbatimEnvironment{promptcode}{Verbatim}{fontsize=\scriptsize, breaklines,
  breakanywhere, breaksymbolleft={}, xleftmargin=0pt}
\newcommand{\pf}[1]{\texttt{\detokenize{#1}}}
\newcommand{\pfb}[1]{\textbf{\pf{#1}}}

\definecolor{cvprblue}{rgb}{0.21,0.49,0.74}
\usepackage[breaklinks,colorlinks,allcolors=cvprblue]{hyperref}
\title{ScopeAct: Scoping Decision Models for Reliable Multi-Step Execution in Mobile GUI Agents}
\author{Linghua Zhang}
\hypersetup{pdftitle={ScopeAct: Scoping Decision Models for Reliable Multi-Step Execution in Mobile GUI Agents}, pdfauthor={Linghua Zhang}}

\begin{document}
\maketitle
\begin{abstract}
Vision-language models (VLMs) enable mobile GUI agents to execute complex tasks, but frequent VLM invocations incur substantial latency and API costs. Prior work mitigates this overhead by delegating low-level action selection to lightweight decision models that choose among executable candidates. However, candidate selection alone lacks the task-level control needed to prevent premature termination and recover from failures, while candidates derived from accessibility trees (AXTrees) may not cover all actions required for task completion. To address these challenges, we introduce \textbf{ScopeAct}. This training-free mobile GUI agent framework combines frequent action selection with sparse VLM reasoning through the Delegation Contract and the Candidate Actions Builder. For task-level control, the Delegation Contract enables the VLM to plan subgoals with explicit completion criteria, recover from failures, and verify task completion. The decision model repeatedly selects actions to execute each subgoal without per-step VLM intervention, returning control to the VLM when execution stalls or final verification is required. For broader action coverage, the Candidate Actions Builder supplements AXTree-derived and system actions with VLM-provided text bindings for task-specific text entry and coordinates for widget-free interactions. Experiments on AndroidWorld demonstrate that ScopeAct achieves 1.7-4.8$\times$ faster execution and 3.7-8.3$\times$ lower API costs than six API-based baselines. ScopeAct requires only 5.3 and 4.6 VLM calls per task under the extended and official evaluation settings, respectively. Under the same settings, it attains 96.3\% and 81.0\% task success rates, surpassing the strongest baselines by 3.4 and 1.7 percentage points, respectively.
\end{abstract}    
\section{Introduction}
\label{sec:intro}

\begin{figure}[!t]
    \centering
    \includegraphics[width=\linewidth]{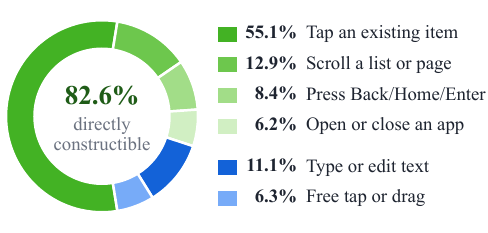}
  \caption{Distribution of 14,823 GUI operations across
812 trajectories from seven AndroidWorld baselines,
including failures.
Green segments (82.6\%) indicate operations directly
constructible from accessibility trees and system actions.
The remaining operations involve text input or editing
(11.1\%) and widget-free taps or drags (6.3\%).
See Appendix~\ref{app:categories} for category definitions.}
    \label{fig:operation-types}
\end{figure}

Recent advances in large language models (LLMs) and
vision-language models (VLMs) have enabled mobile
graphical user interface (GUI) agents to execute
complex, multi-step tasks across smartphone
applications~\cite{zhang2025appagent,wen2024autodroid,
wang2024mobileagentv2,wang2025mobileagente}.
However, frequent VLM calls incur substantial
latency and application programming interface
(API) costs~\cite{yi2026ecoagent,
yu2026graphpilot,chen2025spabench}.
Lightweight decision models offer a promising
alternative by selecting among executable actions
at lower inference cost.
In mobile environments, accessibility trees (AXTrees),
hierarchical representations of the user interface,
expose UI elements and their supported operations,
enabling the construction of executable action
candidates at each step.
Selecting among these candidates can reduce
the cumulative inference overhead of GUI execution.

Prior work has improved GUI action-selection efficiency
through trained selectors based on action verification,
lightweight action transformers, or embedding
prediction~\cite{dai2026vdroid,christianos2025limac,wang2026guijepa}.
These approaches generally select actions for the
entire task, with some relying on generative models
for text entry at execution
time~\cite{dai2026vdroid,christianos2025limac}.
Building on this direction, ScopeAct further improves
efficiency without additional training by delegating
repeated action selection within VLM-defined subgoals
to a decision model, without requiring action parameter
generation during selection.
However, efficient selection alone is insufficient
for reliable GUI execution, which also requires
task-level planning, recovery, and
verification~\cite{ling2026faster}, as well as
adequate coverage of executable actions.
These considerations motivate two key requirements:
task-level control and sufficient action coverage.

The first is task-level control.
A decision model restricted to predefined candidates
cannot independently manage task-level planning,
recover from unexpected states, or verify completion.
A higher-level mechanism is therefore needed to
formulate subgoals, handle execution failures,
and verify task outcomes.
Beyond these responsibilities, the mechanism must
also carry forward information extracted from
the interface to support subsequent actions. 
The second is action coverage.
Candidates derived solely from AXTrees and system
actions cannot cover task-specific text input or
widget-free interactions.
Our analysis of 14,823 GUI operations across
812 trajectories from seven AndroidWorld
baselines~\cite{rawles2025androidworld} shows that
82.6\% are directly constructible without task-specific
values, while 11.1\% involve text input or editing
and 6.3\% require widget-free taps or drags
(\cref{fig:operation-types}).
These findings motivate augmenting candidates
with task-specific text values and coordinates.

\begin{figure*}[t]
    \centering
    \includegraphics[width=\linewidth]{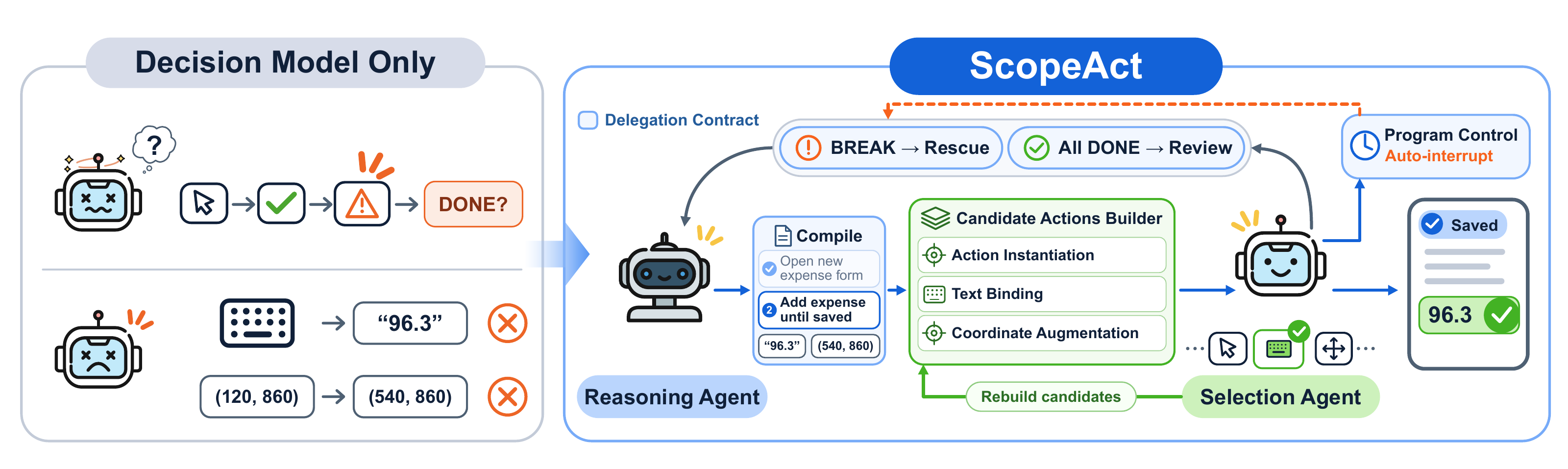}
    \caption{Overview of ScopeAct, compared with a standalone
    decision model (left).
    The Delegation Contract (blue) comprises a Reasoning
    Agent (Compile, Rescue, and Review), Program Control,
    and a Selection Agent.
    The Candidate Actions Builder (green) constructs
    executable candidates from accessibility trees, system
    actions, and task-specific text and coordinate options.}
    \label{fig:overview}
\end{figure*}

To address these challenges, we introduce \textbf{ScopeAct},
a training-free mobile GUI agent framework that combines
decision-model action selection with sparse VLM reasoning,
as illustrated in \cref{fig:overview}.
Its \emph{Delegation Contract} assigns task-level reasoning
to a VLM-based Reasoning Agent and low-level action
selection to a decision-model-based Selection Agent.
The Reasoning Agent decomposes tasks into subgoals
with explicit completion criteria, while the Selection
Agent selects among fully specified candidates within
each subgoal, without generating action parameters.
During execution, Program Control monitors progress,
validates completion signals, and interrupts stalled
or overlong delegations.
Such interruptions, or completion of all subgoals,
return control to the Reasoning Agent for recovery
or review, respectively.
To address the second requirement, the
\emph{Candidate Actions Builder} constructs executable
candidates from AXTrees and system actions, augmented
with VLM-provided text bindings and coordinates for
widget-free taps and drags.
Together, these components enable sustained action
selection across subgoals with VLM guidance when needed.

We evaluate ScopeAct against ten baselines
on AndroidWorld under the extended and official
evaluation settings.
ScopeAct achieves 1.7-4.8$\times$ faster execution
and 3.7-8.3$\times$ lower API costs than six
competitive API-based baselines, requiring only
5.3 and 4.6 VLM calls per task under the
extended and official settings, respectively.
Meanwhile, ScopeAct achieves 96.3\% and 81.0\%
task success in these settings, surpassing
the strongest baselines by 3.4 and 1.7
percentage points, respectively.
Our main contributions are summarized as follows:
\begin{itemize}
    \item We identify task-level control and action
coverage as key requirements for reliable GUI
execution with decision models, quantify coverage
gaps in baseline trajectories, and propose
\textbf{ScopeAct} to address both through its
Delegation Contract and Candidate Actions Builder.

    \item We improve GUI execution efficiency through
training-free, subgoal-conditioned action selection
over preconstructed candidates, reducing the
frequency of costly VLM calls.
    
    \item Experiments on AndroidWorld show that ScopeAct
outperforms all baselines in task success while
improving efficiency over competitive API-based agents.

\end{itemize}

\section{Related work}
\label{sec:related}

\paralead{LLM-based mobile agents.}%
LLMs and VLMs enable mobile agents to execute
natural-language tasks through GUI interactions.
Building on this capability, existing approaches improve execution through
application exploration, experience accumulation,
and memory-based guidance
\cite{mi2026darwinian,
jin2026sega,qin2026eam}.
Others train GUI-specialized models for grounding
and action generation
\cite{hong2024cogagent,cheng2024seeclick,
qin2025uitars,ye2025mobileagentv3},
or generate executable scripts to reduce repeated
inference \cite{wen2025autodroidv2}.
ScopeAct reserves the large model for
task-level decisions and delegates local action
selection to a decision model.

\paralead{Hierarchical and collaborative agents.}%
Hierarchical agents organize planning, execution,
and verification into distinct roles
\cite{agashe2025agents2,
kang2026longhorizonui,han2026vlaagui,
yang2026stateact}.
Hi-Agent jointly optimizes high-level reasoning and
low-level action models~\cite{wu2025hiagent}, while
SWIRL alternates training between planning and interaction
agents~\cite{lu2025swirl}.
In contrast, ScopeAct enables efficient action
selection and subgoal transitions through
training-free delegation with sparse VLM intervention. Collaboration between large and small models combines
high-level reasoning with specialized executors
\cite{gou2025uground,wang2024ponder,
kong2025mobileworld,li2026goclick,
wu2026k2agent,wang2026lamo},
while other methods explore on-device and cloud
collaboration
\cite{zhou2025maiui,xu2026mobileagentv35,
jiang2025openphone}.
CORE uses a local model to rank interface regions
for cloud-based action selection~\cite{fan2025core}.
AdecPilot gives an edge team autonomous tactical
planning and self-correction under cloud-generated
milestones~\cite{li2026adecpilot}.
ScopeAct delegates execution through a finite set of
fully specified candidates, with explicit signals for
subgoal completion and return to the VLM.

\paralead{Candidate-based action selection.}%
V-Droid reduces inference overhead by using a
generative verifier to score discretized action
candidates~\cite{dai2026vdroid}.
LiMAC combines a lightweight Action Transformer
with VLM assistance~\cite{christianos2025limac},
while GUI-JEPA scores executable candidates without
autoregressive decoding~\cite{wang2026guijepa}.
These methods demonstrate the role of candidate
selection in efficient GUI control.
An empirical study of decision models in agent workflows
finds that faster decisions can coexist with reduced
task success and failed recovery~\cite{ling2026faster}.
These findings suggest that efficient candidate
selection alone does not guarantee reliable
multi-step execution.
ScopeAct addresses this limitation through a
Delegation Contract that coordinates VLM-guided
planning, recovery, and verification with
decision-model-based execution.
Its Candidate Actions Builder further addresses
the limitations of predefined action candidates
through task-specific text bindings and
coordinate-based actions.

\section{ScopeAct}
\label{sec:method}

ScopeAct separates task-level VLM reasoning from
low-level action selection by a decision model.
We first describe the execution loop, followed by
the Delegation Contract and Candidate Actions Builder.
\subsection{Framework}
\label{sec:method-overview}

Given a natural-language instruction $u$, a mobile GUI agent aims to complete the specified task through a
sequence of interactions with the device.
At each decision step $t=0,1,\ldots$, the agent receives an observation
$o_t=(I_t,T_t,z_t)$, where $I_t$ denotes the screenshot,
$T_t$ the accessibility tree, and $z_t$ the device context,
including the foreground application and keyboard state.
The agent interacts with the environment through primitive
operations such as tapping, swiping, text entry, and
system navigation.

As illustrated in \cref{fig:overview}, ScopeAct pairs a
Reasoning Agent, implemented by a large VLM $\mathcal{R}$,
with a Selection Agent, implemented by a decision model
$\mathcal{D}$.
The Reasoning Agent plans the task, handles recovery,
and reviews completion before deciding whether to terminate.
Under delegated subgoals, the Selection Agent chooses
among executable action candidates.

ScopeAct coordinates the two agents
through the Delegation Contract and Candidate Actions Builder.
The contract specifies the subgoals assigned to the Selection
Agent and the conditions for transferring control between
the agents.
Using the current observation and information supplied by
the Reasoning Agent, the builder constructs the executable
candidates available at each step.
The program validates selected operations before executing
them and updates the observation for the next decision.

Execution begins with the Reasoning Agent compiling
the instruction into an ordered sequence of subgoals,
each specifying a local objective, its completion condition,
and any text values required for execution.
Within each subgoal, the Selection Agent repeatedly
selects among fully specified candidates supplied
by the builder, without generating new actions
or parameters during selection.
When the program accepts a subgoal completion signal,
execution advances directly to the next planned subgoal
if one remains, without an additional VLM call.

Control returns to the Reasoning Agent for Rescue
when the Selection Agent cannot proceed or Program
Control detects stalled or prolonged execution as shown in \cref{fig:overview}.
Completion of all planned subgoals instead triggers
Review. Following either intervention, the Reasoning Agent can
revise the remaining plan and resume delegation, or
terminate the task if it judges that the instruction
has been fulfilled.
Execution also ends if the Reasoning Agent reports that
the task is blocked or the execution budget is exhausted.
The episode in Figure~\ref{fig:case-study} of
Appendix~\ref{app:case-study} follows this control flow
from planning to final review.

\subsection{Delegation Contract}
\label{sec:delegation}

The Delegation Contract specifies the subgoal
information passed to the Selection Agent and
the conditions for returning control to the
Reasoning Agent.
The Selection Agent selects executable actions
and signals completion or interruption, while
the Reasoning Agent supervises execution through
Compile, Rescue, and Review.

\paralead{Subgoal compilation.}%
At the beginning of each task, the Reasoning Agent
compiles the natural-language instruction and initial
device observation into an ordered sequence of
subgoals and accompanying semantic notes, which
together guide subsequent task execution:
\begin{equation}
    \label{eq:subgoal-compilation}
    (G,N)=\mathcal{P}_{\mathrm{compile}}(u,o_0),
\end{equation}
where $G=(g_1,\ldots,g_K)$ is a sequence of $K$ subgoals
and $N$ stores semantic notes needed across execution steps.
The Reasoning Agent can update these notes in subsequent
calls to preserve information retrieved from other applications and current screen.

For subgoal $k\in\{1,\ldots,K\}$ in the current plan, we represent its
objective and text requirements by $g_k=(q_k,B_k)$.
Here, $q_k$ specifies a local objective and its
completion condition, such as ``stop when the saved
expense is displayed''.
The set $B_k=\{(f_j,v_j)\}_{j=1}^{J_k}$ contains $J_k$
text bindings, each pairing a field description $f_j$
with the exact text value $v_j$ to enter,
such as ``Amount: 307.01'' in
Figure~\ref{fig:case-study}.
A subgoal that requires no text entry has $B_k=\varnothing$.
Beyond text bindings, the Reasoning Agent can also
provide coordinate-based action options for each subgoal and update both as execution proceeds, as described in
\cref{sec:candidates}.
A single compilation can produce multiple subgoals,
which the Selection Agent executes sequentially.
For each subgoal, we define a delegation as an
uninterrupted period of action selection.

\paralead{Action selection and control transfer.}%
Given a delegated subgoal $g_k$, the Candidate Actions
Builder constructs a set $\mathcal{C}_t$ of executable
candidates from the current observation and delegated
information, as detailed in \cref{sec:candidates}.
These candidates define the available actions for
the Selection Agent at each step.
To select among them, the Selection Agent receives
the core decision context
$s_t=(g_k,h_{k,t},I_t,T_t)$,
where $h_{k,t}$ records the actions executed so far
within the current delegation.
The request also includes the observation identifier
and any selections rejected by program control,
as specified in Appendix~\ref{app:prompt-luna}.
Each candidate is presented with its identifier and
semantic description, enabling the Selection Agent
to choose directly among executable actions.
Based on this context and candidate set, the decision
model $\mathcal{D}$ selects an executable candidate
The decision model $\mathcal{D}$ chooses a candidate identifier by its identifier or returns one of two control signals:, summarized by the output $y_t$:
\begin{equation}
  \label{eq:action-selection}
  \begin{aligned}
    y_t &= \mathcal{D}(s_t,\mathcal{C}_t),\\
    y_t &\in \operatorname{ID}(\mathcal{C}_t)
        \cup \{\texttt{DONE},\texttt{BREAK}\},
  \end{aligned}
\end{equation}
where $\operatorname{ID}(\mathcal{C}_t)$ denotes the
identifiers of the available executable candidates.
When $y_t$ is a candidate identifier, the program
validates the corresponding operation before execution.
After executing an accepted operation, it updates the observation
and records execution feedback for the next decision.

A \texttt{DONE} output reports that the completion condition
in $q_k$ is met.
Before accepting the signal, the program applies progress
checks, including whether required bound values have been entered.
If another planned subgoal remains, an accepted \texttt{DONE}
advances execution to that subgoal without invoking the
Reasoning Agent.
Otherwise, it triggers Review.
By returning \texttt{BREAK}, the Selection Agent requests
Rescue when execution cannot proceed reliably.
The program also tracks progress and can interrupt execution
upon repeated screens, lack of progress, or blocked targets.
A interruption is triggered when a delegation reaches
its limit of 12 actions.
We refer to the program's progress tracking and automatic
interruption rules as Program Control.
Both \texttt{BREAK} and program-triggered interruptions return
control to the Reasoning Agent for Rescue.

\paralead{Rescue.}%
When execution is interrupted, the Reasoning Agent
examines the current observation, execution history,
and interruption reason.
Based on this assessment, it determines the appropriate
recovery strategy, including revising the current
subgoal and its text bindings, replanning subsequent
subgoals, or providing coordinate-based actions
for widget-free interactions.
The subsequent control decision depends on task progress.
Verified task completion leads to termination,
subgoal completion advances execution to the next
subgoal, and unfinished work resumes under the
revised delegation.

\paralead{Review.}%
Once the planned subgoal sequence is exhausted,
the Reasoning Agent reviews the resulting device state
against the original instruction $u$ and accumulated
semantic notes $N$.
Based on this review, unmet requirements trigger
updates to the notes and delegation of additional
subgoals to the Selection Agent.
Successful verification instead leads to task
termination and a final result with supporting evidence.

\subsection{Candidate Actions Builder}
\label{sec:candidates}

The Candidate Actions Builder transforms the current
observation and delegated information into the finite
set $\mathcal{C}_t=\{c_t^1,c_t^2,\ldots,c_t^{M_t}\}$,
where $M_t$ is the number of executable candidates at step $t$,
and $c_t^i$ is the $i$th candidate for $i=1,\ldots,M_t$.
Each candidate contains a unique identifier, a
semantic description, and a fully specified executable
operation.
Given the current observation and any text bindings or
coordinate-based actions supplied by the Reasoning Agent,
the builder constructs candidates deterministically.

\paralead{Action instantiation.}%
The builder traverses the accessibility tree to
identify visible, enabled, and interactable elements.
For each element, it instantiates the actions that the
element's supported operations and current state allow,
including taps, long presses, and scrolls.
Editable fields receive text-entry candidates for bound values,
and sliders receive adjustment candidates.
It also includes system-level actions such as Back,
Home, and application launching.
The current observation provides grounding parameters,
such as element coordinates and available scrolling directions.
To distinguish similar controls, candidate
descriptions incorporate interface context,
including element text, content descriptions,
and surrounding labels.

\paralead{Text binding.}%
Text-entry candidates combine editable fields from the
accessibility tree with the text bindings $B_k$ supplied
by the Reasoning Agent.
To construct these candidates, the builder matches each
binding to input fields based on their labels and
surrounding context.
For example, pairing the field description \texttt{Amount}
with the value \texttt{307.01} produces a candidate such as
\texttt{Enter "307.01" in field "Amount"}.
Ambiguous matches are retained as separate candidates,
allowing the Selection Agent to choose the intended target.
Once selected, the program enters the bound value
into the corresponding field.

\paralead{Coordinate-based augmentation.}%
For widget-free taps and drags, the
Reasoning Agent can supply coordinate-based actions when
delegating subgoals during.
The builder incorporates these actions into
$\mathcal{C}_t$ as executable candidates alongside
accessibility-derived actions.
The Selection Agent chooses among the augmented candidates
through the same selection mechanism.

Candidate identifiers correspond to executable operations
defined for the current observation and are valid only
within that decision step.
To keep the candidates consistent with the evolving
interface, the program rebuilds the candidate set
after each environment interaction using the updated
observation for the next decision.

\begin{table*}[t]
    \centering
    \small
    \setlength{\tabcolsep}{3pt}
    \renewcommand{\arraystretch}{0.95}
    \caption{Task success by AndroidWorld difficulty level
under the extended (Ext.) and official (Off.) budgets.
Success rates (\%) are the mean $\pm$ sample standard
deviation across three task seeds, each containing
61 Easy, 36 Medium, and 19 Hard instances.
Evaluation follows \cref{sec:setup}, and bold
marks the highest mean in each column.}
    \label{tab:difficulty}
    \begin{tabular*}{1\linewidth}
    {@{\extracolsep{\fill}}lcccccc@{}}
        \toprule
        Method
        & Easy (Ext.) & Medium (Ext.) & Hard (Ext.)
        & Easy (Off.) & Medium (Off.) & Hard (Off.) \\
        \midrule
        SeeAct-V
        & 92.3\pmstd{1.9}
        & 90.7\pmstd{4.2}
        & 68.4\pmstd{5.3}
        & 85.2\pmstd{3.3}
        & 65.7\pmstd{1.6}
        & 59.6\pmstd{8.0} \\
        MobileWorld PE
        & 95.6\pmstd{0.9}
        & \textbf{95.4}\pmstd{3.2}
        & 70.2\pmstd{8.0}
        & 76.5\pmstd{2.5}
        & 50.9\pmstd{4.2}
        & 42.1\pmstd{5.3} \\
        M3A
        & 92.9\pmstd{3.8}
        & 90.7\pmstd{7.0}
        & 71.9\pmstd{3.0}
        & 89.6\pmstd{2.5}
        & \textbf{75.9}\pmstd{10.5}
        & 52.6\pmstd{5.3} \\
        T3A
        & 88.5\pmstd{0.0}
        & 88.9\pmstd{4.8}
        & 68.4\pmstd{0.0}
        & 82.0\pmstd{1.6}
        & 67.6\pmstd{10.5}
        & 49.1\pmstd{8.0} \\
        Minitap
        & 97.3\pmstd{1.9}
        & 91.7\pmstd{5.6}
        & 80.7\pmstd{3.0}
        & 89.1\pmstd{0.9}
        & 66.7\pmstd{7.3}
        & 61.4\pmstd{3.0} \\
        MobileUse
        & 94.0\pmstd{0.9}
        & 82.4\pmstd{4.2}
        & 52.6\pmstd{5.3}
        & 86.3\pmstd{3.8}
        & 63.0\pmstd{5.8}
        & 31.6\pmstd{5.3} \\
        GUI-Owl-7B
        & 73.8\pmstd{4.9}
        & 48.1\pmstd{7.0}
        & 14.0\pmstd{6.1}
        & 72.1\pmstd{6.6}
        & 47.2\pmstd{7.3}
        & 12.3\pmstd{8.0} \\
        MAI-UI-8B
        & 81.4\pmstd{3.4}
        & 60.2\pmstd{8.9}
        & 40.4\pmstd{6.1}
        & 71.0\pmstd{2.5}
        & 45.4\pmstd{5.8}
        & 28.1\pmstd{8.0} \\
        V-Droid
        & 65.0\pmstd{5.0}
        & 42.6\pmstd{8.9}
        & 28.1\pmstd{12.2}
        & 62.3\pmstd{6.6}
        & 34.3\pmstd{6.4}
        & 14.0\pmstd{6.1} \\
        EcoAgent
        & 46.4\pmstd{3.4}
        & 32.4\pmstd{5.8}
        & 12.3\pmstd{6.1}
        & 27.9\pmstd{4.3}
        & 10.2\pmstd{5.8}
        & 1.8\pmstd{3.0} \\
        \midrule
        ScopeAct
        & \textbf{97.8}\pmstd{0.9}
        & \textbf{95.4}\pmstd{1.6}
        & \textbf{93.0}\pmstd{8.0}
        & \textbf{92.3}\pmstd{3.4}
        & 66.7\pmstd{4.8}
        & \textbf{71.9}\pmstd{6.1} \\
        \bottomrule
    \end{tabular*}
\end{table*}

\begin{figure*}[t]
    \centering
    \includegraphics[width=\linewidth]
    {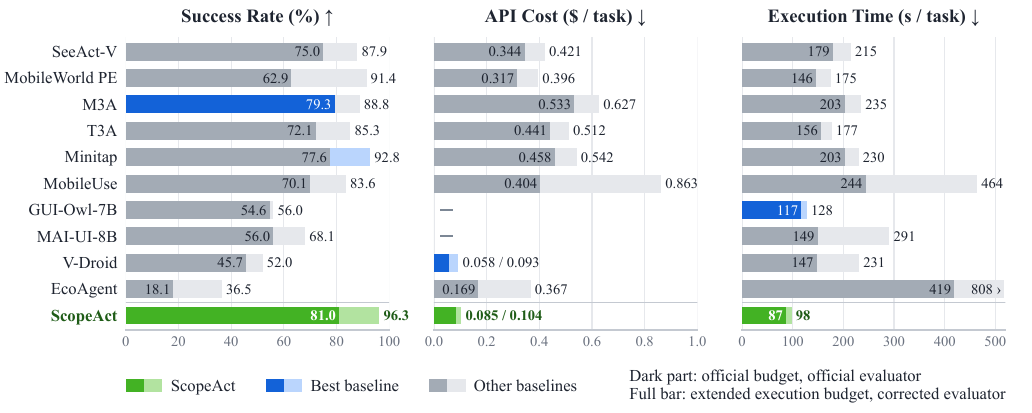}
    \caption{Task success and execution efficiency
on AndroidWorld for ScopeAct and ten baselines, averaged
across three task seeds with 348 instances per method.
Full bars show extended-budget results and dark segments
show official-budget results under the protocols in \cref{sec:setup},
with failed runs included in per-task API cost and execution time.
Green denotes ScopeAct and blue denotes the best
baseline for each metric and evaluation setting,
with self-hosted models excluded from the API-cost ranking
because GPU costs are not included.}
    \label{fig:main-results}
\end{figure*}

\section{Experiments}
\label{sec:experiments}

Our experiments compare ScopeAct against ten
baselines on AndroidWorld, examining task success
and execution efficiency under an extended execution
budget and at the official action limit.
Both settings include aggregate results and a
breakdown by task difficulty.

\subsection{Evaluation setup}
\label{sec:setup}

We evaluate ScopeAct against ten baselines on
AndroidWorld, which contains 116 task templates
across 20 applications and evaluates success
based on the resulting device state.
Three task seeds (30, 42, and 20261008) yield
348 instances per method.
The baselines span five categories:
large-small model systems (SeeAct-V~\cite{gou2025uground}
and MobileWorld PE~\cite{kong2025mobileworld}),
single-model agents (M3A and T3A~\cite{rawles2025androidworld}),
multi-agent systems (Minitap~\cite{favreau2026minitap}
and MobileUse~\cite{li2025mobileuse}),
specialized GUI models (GUI-Owl-7B~\cite{ye2025mobileagentv3}
and MAI-UI-8B~\cite{zhou2025maiui}),
and candidates selection agents (V-Droid and EcoAgent).
For consistency, all large-model roles use GPT-6.1 Sol,
while ScopeAct uses the GPT-6 Luna Decisions API
for its Selection Agent.
API request settings are detailed in
Appendix~\ref{app:baselines},
Table~\ref{tab:api-params}.

We evaluate under two settings.
The extended execution budget allows up to 1800\,s and 100 environment
actions per task and scores success with a corrected evaluator.
The official budget applies AndroidWorld's complexity-based action limit,
with a median of 12 model actions, retrospectively to the same
trajectories and keeps only the successes that the original evaluator
accepts within the limit.
Appendix~\ref{app:budgets} defines both protocols.
Performance is assessed using four primary metrics:
Extended Success ($\uparrow$), Official Success
($\uparrow$), API Cost ($\downarrow$), and
Execution Time ($\downarrow$).
The first two measure success under the extended budget
and retained success at the official action limit,
respectively.
API Cost is the model API charge per task, while
Execution Time is the online wall-clock time excluding
initialization and evaluation.
Self-hosted models incur zero reported API charges,
excluding GPU costs.
To further assess decision efficiency, we report
completed Sol API requests as VLM calls and count
executed model actions.
Results are averaged equally across three seeds with sample standard deviations.
Appendix~\ref{app:evaluation} details model
configurations, API accounting, and evaluator corrections.

\subsection{Extended execution budget}
\label{sec:relaxed}

\Cref{fig:main-results} compares task success and
efficiency under both evaluation settings, with
full results in Appendix~\ref{app:full-results}.
Under the extended execution budget, ScopeAct
achieves 96.3\% mean task success, outperforming
all ten baselines, including the strongest
competitors, Minitap (92.8\%) and
MobileWorld PE (91.4\%).

ScopeAct averages \$0.104 in API cost and
98\,s in execution time per task.
In comparison, the six API-based baselines with over
80\% task success require \$0.396--0.863 and 175--464\,s,
corresponding to cost reductions of 3.8--8.3$\times$
and speedups of 1.8--4.8$\times$, respectively.
Among the remaining baselines, V-Droid achieves a
slightly lower API cost of \$0.093 but only 52.0\%
task success, 44.3 percentage points below ScopeAct.
EcoAgent, by contrast, requires 808\,s per task,
over 8$\times$ ScopeAct's execution time.

At the model-call level, ScopeAct requires only
5.3 VLM calls per task, compared with 18.7--85.0
for the six competitive API-based baselines.
Meanwhile, its decision model receives 18.0 requests
per task.
Despite fewer VLM calls, ScopeAct maintains comparable
trajectory lengths, with 14.3 model actions per task,
within the 13.3--14.6 range of M3A, T3A, Minitap,
and SeeAct-V, as shown in Table~\ref{tab:relaxed}
of the supplementary material.
Together, these call and action counts indicate that
ScopeAct reduces VLM involvement while maintaining
comparable trajectory lengths.

To assess performance across task difficulties, we
group instances by AndroidWorld's difficulty labels,
distinct from the complexity values used for action limits.
\Cref{tab:difficulty} reports success under both
evaluation settings.
Under the extended budget, ScopeAct achieves
97.8\%, 95.4\%, and 93.0\% success on Easy,
Medium, and Hard tasks, respectively, leading
on Easy and Hard while matching MobileWorld PE
on Medium.
Notably, its success drops by only 4.8 percentage
points from Easy to Hard, compared with at least
16.6 points for all baselines.
On Hard tasks, ScopeAct further outperforms
Minitap by 12.3 points and uniquely solves
eight instances versus one for Minitap,
demonstrating its robustness on challenging tasks.

\subsection{Official budget}
\label{sec:model-action-cap}

At the official action limit, ScopeAct
retains 81.0\% task success,
the highest mean among the evaluated methods, ahead of
M3A at 79.3\% and Minitap at 77.6\%.
In terms of efficiency, it also achieves the shortest
execution time among all evaluated methods,
averaging 87\,s per task with only
4.6 VLM calls and an API cost of
\$0.085.
By comparison, the six API-based baselines with over 80\%
success under the extended budget require
\$0.317--0.533 and 146--244\,s per task
under the official budget.
Their API costs are 3.7--6.3 times ScopeAct's, and their
execution times are 1.7--2.8 times as long.
Among the remaining baselines, V-Droid achieves a lower
API cost of \$0.058 but only 45.7\%
task success.

The difficulty breakdown in \cref{tab:difficulty}
shows that ScopeAct leads on Easy and Hard tasks
at the official action limit, achieving 92.3\%
and 71.9\% success, respectively.
On Medium tasks, however, its 66.7\% success
trails M3A's 75.9\%.
For ScopeAct, the median action-budget usage
is higher on Medium than Hard tasks
(75\% vs. 64\%).
This pattern is consistent with greater sensitivity
to the action limit on Medium tasks.
The impact of the evaluation setting is also
evident in MobileWorld PE, whose success drops
from 91.4\% under the extended budget to
62.9\% under the official budget.
Further results and budget statistics are
provided in Appendix~\ref{app:full-results}.

\section{Ablation study}
\label{sec:ablation}

Three groups of ablations examine the contributions
of ScopeAct's framework components, their internal
mechanisms, and the models underlying its agent roles.
Following the main evaluation protocol, each variant
is evaluated once on each of the three task seeds,
30, 42, and 20261008.
We report extended-budget success and success retained
at the official action limit using the protocol in
\cref{sec:setup}, with API cost, execution time, and
VLM calls measured under the extended budget.
Across all variants, success rates are reported as
mean $\pm$ sample standard deviation across seeds,
while efficiency values are per-task means including
failures.
Further details on the ablation design and evaluation
protocol are provided in Appendix~\ref{app:ablation},
with additional statistics in
Appendices~\ref{app:ablation-stats} and~\ref{app:coordinate-tasks}.

\subsection{Framework-level ablations}
\label{sec:ablation-modules}

At the component level, the Delegation Contract and
Candidate Actions Builder are ablated separately,
with the remaining framework unchanged.
To isolate their effects, we remove each component
while retaining the remaining framework.
Without delegation, Sol replaces the Selection Agent
and selects actions from the same builder at every step.
Without the builder, candidates enumerate visible
accessibility-tree nodes with tap, long-press, and
swipe actions for each node.
\Cref{tab:ablation} reports the results.

\begin{table}[t]
    \centering
    \small
    \setlength{\tabcolsep}{1pt}
    \caption{Framework ablations, removing either the
Delegation Contract or the Candidate Actions Builder.
Success rates (\%) are means $\pm$ sample standard
deviations over three seeds of 116 instances each, with
bold marking the highest mean in each success column.
VLM calls, API cost (\$), and time (s) are per-task
means under the extended budget, including failures.}
    \label{tab:ablation}
    \begin{tabular*}{1\linewidth}
        {@{\extracolsep{\fill}}lccccc@{}}
        \toprule
        Variant & Extended & Official &
        VLM calls & API cost & Time \\
        \midrule
        \textbf{ScopeAct}
        & \textbf{96.3}\pmstd{1.3}
        & \textbf{81.0}\pmstd{2.3}
        & 5.3 & 0.104 & 98 \\
        \ablrow w/o delegation
        & 79.9\pmstd{0.5}
        & 72.4\pmstd{0.9}
        & 20.0 & 0.600 & 146 \\
        \ablrow w/o builder
        & 58.0\pmstd{1.8}
        & 46.6\pmstd{1.7}
        & 8.8 & 0.201 & 129 \\
        \bottomrule
    \end{tabular*}
\end{table}

Without the Delegation Contract, task success drops
from 96.3\% to 79.9\% under the extended budget
and from 81.0\% to 72.4\% under the official budget.
This degradation is accompanied by substantial
efficiency costs: VLM calls increase from 5.3 to
20.0 per task, API cost rises by 5.8$\times$,
and execution time increases from 98\,s to 146\,s.
However, on the 273 instances solved by variant and ScopeAct,
model action counts remain comparable at 11.3
for ScopeAct and 10.9 without delegation.
This suggests that the efficiency gains primarily
come from reduced VLM involvement.

Excluding the Candidate Actions Builder yields
58.0\% and 46.6\% success, respectively,
lower than variants without Text Binding or
coordinate actions alone in \cref{tab:ablation-internal}.
On the 199 instances solved by variant and ScopeAct,
model actions increase from 9.0 to 11.4 per task.
Across all instances, Rescue calls increase from
1.20 to 4.15 per task, while VLM calls rise from
5.3 to 8.8 and API cost increases from \$0.104 to \$0.201.
These results support the combined role of semantic
candidate construction, text bindings, and coordinate
actions in efficient execution and recovery.

\subsection{Mechanism ablations}
\label{sec:ablation-delegation}

The mechanism-level analysis separately ablates
Rescue, Review, and Program Control within the
Delegation Contract.

The Rescue ablation terminates execution at the first
interruption, whereas the Review ablation ends it
once all planned subgoals are marked complete.
Disabling Program Control removes automatic
interruptions for repeated screens, stalled progress,
or blocked targets, along with the 12-action
delegation limit.
For the Candidate Actions Builder, we ablate
Text Binding and coordinate-based augmentation.
In the Text Binding ablation, values are entered
through on-screen keyboard actions, while the
coordinate ablation excludes coordinate-based
actions provided by the Reasoning Agent.
\Cref{tab:ablation-internal} reports the results.

\begin{table}[t]
    \centering
    \small
    \setlength{\tabcolsep}{1pt}
    \caption{Mechanism ablations for the Delegation
Contract above the separator and the Candidate Actions Builder below it.
Metrics and reporting conventions follow \cref{tab:ablation}.}
    \label{tab:ablation-internal}
    \begin{tabular*}{1\linewidth}
        {@{\extracolsep{\fill}}lccccc@{}}
        \toprule
        Variant & Extended & Official &
        VLM calls & API cost & Time \\
        \midrule
        \textbf{ScopeAct}
        & \textbf{96.3}\pmstd{1.3}
        & \textbf{81.0}\pmstd{2.3}
        & 5.3 & 0.104 & 98 \\
        \ablrow w/o rescue
        & 60.3\pmstd{2.6}
        & 58.0\pmstd{1.3}
        & 2.8 & 0.045 & 57 \\
        \ablrow w/o review
        & 74.4\pmstd{3.9}
        & 64.7\pmstd{2.3}
        & 3.2 & 0.060 & 75 \\
        \ablrow w/o prog.\ control
        & 95.1\pmstd{1.3}
        & 77.3\pmstd{3.3}
        & 5.4 & 0.110 & 104 \\
        \midrule
        \ablrow w/o text binding
        & 77.6\pmstd{0.9}
        & 49.7\pmstd{0.5}
        & 8.2 & 0.258 & 187 \\
        \ablrow w/o coordinates
        & 93.4\pmstd{1.0}
        & 74.1\pmstd{2.3}
        & 5.3 & 0.130 & 100 \\
        \bottomrule
    \end{tabular*}
\end{table}

Rescue and Review address distinct limitations
of delegated execution.
Without Rescue, 150 instances terminate at their
first execution interruption, reducing success
to 60.3\% and 58.0\% under the two budgets.
Review, in turn, provides task-level verification
beyond local subgoal completion.
Without Review, 280 instances terminate once their
planned subgoals are exhausted, and success drops
to 74.4\% and 64.7\%.
To further examine the contribution of Review
beyond answer generation,
Appendix~\ref{app:audit-paired}
compares the variants on 254 non-question-answering
instances that exhausted their planned subgoals
without Review.
Among these instances, ScopeAct alone solves 56,
compared with only one for the variant.
This result further highlights the importance of
task-level verification and subsequent correction
beyond local subgoal planning.

Removing Program Control reduces task success
to 95.1\% and 77.3\% under the two budgets.
In the full framework, Program Control triggers
118 of the 414 Rescue calls, while the Selection
Agent's \texttt{BREAK} accounts for the remaining 296.
Without Program Control, model actions increase
from 14.3 to 16.0 per task, and execution time
rises from 98\,s to 104\,s.
These results highlight the importance of proactively
interrupting stalled delegations rather than relying
solely on model-generated \texttt{BREAK} signals.

The Text Binding ablation increases average model
actions from 14.3 to 34.7 per task and execution
time from 98\,s to 187\,s.
This increased overhead is accompanied by a drop
in task success to 77.6\% under the extended budget
and 49.7\% under the official budget.
By incorporating task-specific input values directly
into executable candidates, Text Binding replaces
sequences of keyboard actions with a single selection,
reducing action overhead and improving task success,
particularly under constrained action budgets.

Excluding coordinate-based actions reduces task
success to 93.4\% and 74.1\% under the two budgets.
To examine this drop, we focus on the five
task templates involving drawing or map interactions.
Among the 13 instances solved only by ScopeAct,
10 belong to these templates, suggesting that
much of the success difference stems from tasks
requiring actions beyond accessibility-tree elements.

\subsection{Model ablations}
\label{sec:ablation-decision}

To examine the effects of selector capacity and
task-level reasoning, we evaluate three variants.
To isolate selector capacity, we replace Luna in
the Selection Agent with Sol or Qwen3.5-9B
~\cite{qwen2026qwen35}, denoted Luna$\to$Sol
and Luna$\to$Qwen, respectively.
Qwen is a self-hosted open-weight model, comparable
in parameter count to V-Droid's 8B verifier,
and is used without additional GUI-specific tuning.
Both variants retain the Delegation Contract,
Candidate Actions Builder, and decision inputs.
To assess task-level reasoning, the third variant,
w/o Reasoning, assigns Luna the entire task as a
single subgoal without text bindings or
Reasoning Agent calls.
\Cref{tab:ablation-decision} reports success
for all three variants, along with selector
cost and time under the extended budget.

Using Sol instead of Luna improves success from
96.3\% to 96.8\% under the extended budget and
from 81.0\% to 84.8\% at the official action limit,
but increases selector API cost by 22$\times$
and selector time by 4.3$\times$.
These modest gains suggest that a
decision model can perform low-level action selection
effectively under structured delegation, with
substantially lower overhead.
With Sol held fixed as the selector, removing
delegation reduces success to 79.9\% and 72.4\%
under the two budgets, respectively.
This further highlights the contribution of
structured delegation beyond selector model capacity.

\begin{table}[t]
    \centering
    \small
    \setlength{\tabcolsep}{1pt}
    \caption{Ablations of the selector and Reasoning Agent,
with cost and time measured only for the selector.
Reporting conventions follow \cref{tab:ablation}, with
selector cost and time also given as mean $\pm$ sample
standard deviation across the three seeds.
$^\dagger$Qwen3.5-9B is self-hosted, with zero API
charge and GPU costs excluded.}
    \label{tab:ablation-decision}
    \begin{tabular*}{1\linewidth}
        {@{\extracolsep{\fill}}lcccc@{}}
        \toprule
        Variant & Extended & Official &
        Selector cost & Selector time \\
        \midrule
        \textbf{ScopeAct}
        & 96.3\pmstd{1.3}
        & 81.0\pmstd{2.3}
        & 0.016\pmstd{0.000}
        & 14.5\pmstd{0.4} \\
        \ablrow Luna$\to$Sol
        & \textbf{96.8}\pmstd{1.3}
        & \textbf{84.8}\pmstd{2.0}
        & 0.351\pmstd{0.012}
        & 62.2\pmstd{4.3} \\
        \ablrow Luna$\to$Qwen
        & 89.9\pmstd{1.8}
        & 57.8\pmstd{3.1}
        & 0$^\dagger$
        & 111.1\pmstd{14.1} \\
        \ablrow w/o reasoning
        & 16.7\pmstd{1.3}
        & 16.7\pmstd{1.3}
        & 0.003\pmstd{0.000}
        & 3.3\pmstd{0.2} \\
        \bottomrule
    \end{tabular*}
\end{table}

Replacing Luna with Qwen3.5-9B achieves 89.9\%
success under the extended budget, 6.3 percentage
points below ScopeAct's 96.3\%, demonstrating
the feasibility of a self-hosted selector without
additional tuning.
However, the gap widens at the official action limit,
where Qwen retains 57.8\% success compared with
81.0\% for Luna.
To examine this difference, we compare the 310
instances solved by both variants.
Qwen executes 18.6 model actions per task,
compared with 12.7 for Luna.
Across all instances, Qwen also issues \texttt{BREAK}
less frequently, at 0.02 versus 0.85 times per task.
These results highlight the importance of selector
behavior for both execution efficiency and
task-level control.
Beyond selector choice, we further examine the
importance of task-level reasoning.
Without the Reasoning Agent, success drops to
16.7\% under both budgets, highlighting the
importance of task-level control beyond
low-level action selection.

\section{Conclusion}
\label{sec:conclusion}

We present \textbf{ScopeAct}, a training-free mobile GUI
agent framework that combines lightweight action selection
with sparse VLM reasoning.
Its Delegation Contract assigns task-level planning,
recovery, and verification to the Reasoning Agent,
while the Candidate Actions Builder extends action
coverage through text bindings and coordinate-based actions.
Together, these components enable efficient multi-step
execution with task-level control and broader action coverage.
On AndroidWorld, ScopeAct achieves 96.3\% and 81.0\%
task success under the extended and official budgets,
respectively.
It also achieves 1.8--4.8$\times$ speedups and
3.8--8.3$\times$ lower API costs than six competitive
API-based baselines under the extended budget.
These results demonstrate that lightweight action
selection with sparse task-level VLM supervision
enables efficient and reliable multi-step execution
in mobile GUI environments.

\section*{LLM usage statement}
We used large language models to edit the language of this
manuscript and to help write analysis and plotting scripts.
All ideas, experiments, results, and claims are the authors' own
and were checked by them.
GPT-6.1 Sol and GPT-6 Luna are components of the evaluated
systems, as described in \cref{sec:setup}.

{\small
  \bibliographystyle{ieeenat_fullname}
  \bibliography{references}
}

\clearpage
\setcounter{figure}{0}
\setcounter{table}{0}
\makeatletter
\setlength{\@fptop}{0pt}
\setlength{\@dblfptop}{0pt}
\makeatother
\maketitlesupplementary
\appendix
\renewcommand{\thetable}{S\arabic{table}}
\renewcommand{\thefigure}{S\arabic{figure}}
\renewcommand{\theHtable}{S\arabic{table}}
\renewcommand{\theHfigure}{S\arabic{figure}}
\section{Operation categories}
\label{app:categories}

We analyze 14,823 interface operations from 812
seed-30 trajectories of seven of the ten AndroidWorld
baselines, including failed runs. Among these, 671 successful
trajectories contain 8,278 operations.
SeeAct-V is excluded because its traces lack
accessibility trees. User-facing messages and wait
actions are excluded from the operation counts.

Each operation is matched against the accessibility
tree immediately preceding it and classified into
one of five categories.
The categories distinguish operations whose parameters are available locally from text
entry and interactions outside accessible controls.
\begin{itemize}
  \item \emph{Available-option selection} covers taps and long presses on existing interface
  elements, directional scrolling, and system navigation such as Back, Home, and Enter.
  It also includes launching or closing applications and entering instruction-provided text.
  \item \emph{Other text input} requires values retrieved from elsewhere or composed.
  \item \emph{Keyboard-key taps} target individual keys on the on-screen keyboard.
  \item \emph{Locations without corresponding controls} include coordinates on maps,
  canvases, or images.
  \item \emph{Free gestures} include dragging beyond directional scrolling.
\end{itemize}
Instruction-provided text is identified by matching
input values against the task instruction, ignoring
case and punctuation, and includes recipe titles,
event names, filenames, and contact names.
For Fig.~1 of the main paper, we group instruction-provided text entry, other text input,
and keyboard-key taps into the text-input slice.
Locations without corresponding controls and free gestures form the widget-free slice.
The remaining available-option operations form the green segments, which therefore exclude
instruction-provided text entry.

\section{Evaluation setup and full results}
\label{app:evaluation}

This section specifies the model configurations and evaluation protocol used in the main
experiments.
The full results then report task success and execution efficiency under both budgets.

\subsection{Baselines and models}
\label{app:baselines}

All methods retain their respective control structures, and \cref{tab:relaxed} lists the
models each one uses. All large-model roles use GPT-6.1 Sol through its API, and ScopeAct uses
GPT-6 Luna, also through its API, as its decision model. Their request settings are listed in \cref{tab:api-params}. The small models used by the baselines are served via vLLM~\cite{kwon2023efficient} on an
NVIDIA A100 GPU. V-Droid and EcoAgent run their released code with only the large model replaced.
V-Droid scores candidate actions with its released V-Droid-8B verifier, and EcoAgent executes
with UI-TARS-1.5-7B~\cite{bytedanceseed2025uitars15} and observes with
Qwen2-VL-2B~\cite{wang2024qwen2vl}.
MobileWorld PE uses UI-Ins-7B~\cite{chen2026uiins} as its grounding executor.

\begin{table}[!ht]
  \centering
  \small
  \caption{API request settings used for Sol and Luna.
  Sol requests with function tools use the Responses API with the same settings, and Luna
  returns a listed choice.}
  \label{tab:api-params}
  \begin{tabular*}{1\linewidth}{@{\extracolsep{\fill}}lcc@{}}
    \toprule
    Setting & GPT-6.1 Sol & GPT-6 Luna \\
    \midrule
    Endpoint & Chat Completions & Decisions \\
    Reasoning effort & Medium & Not set \\
    Max output tokens & 16,384 & Not set \\
    Max input tokens & Not set & Not set \\
    Image input & Screenshot & $\le 2^{20}$ pixels \\
    Timeout (s) & 180 & 60 \\
    \bottomrule
  \end{tabular*}
\end{table}

The Luna endpoint restricts its output to the supplied choices and exposes no output-length
setting.
Its input screenshot is downscaled to at most $2^{20}$ pixels before transmission.

\subsection{API cost accounting}
\label{app:api-cost}

Sol costs are computed at the experiment's accounting rates of \$2 per million input tokens,
\$10 per million output tokens, and \$0.10 per million cached input tokens.
Luna costs use the API-reported charge.
For self-hosted inference, the reported API cost is zero and excludes GPU costs.

\subsection{Execution budgets}
\label{app:budgets}

The two evaluation settings differ in their execution budgets and
success evaluators.
The extended execution budget allows up to 1800\,s and 100 environment
actions per task, and its success is judged by the corrected evaluator
described in \cref{app:evaluator}, which also accepts valid end states
that the official checks reject.
By contrast, the official budget follows AndroidWorld's complexity-based
limit of $\lfloor 10c\rfloor$ model actions for a task of complexity $c$,
giving a median limit of 12 over the 116 task templates.
A model action counts one decision executed on the device, even when it
expands into several primitive operations.
We apply this limit retrospectively to the extended-budget trajectories,
so both settings are scored from the same runs.
A trajectory counts as an official success only when the original
AndroidWorld evaluator passes and the trajectory ends within the limit.
Efficiency metrics are computed from the trajectory prefix up to the limit.

\subsection{Corrected evaluator}
\label{app:evaluator}

The official AndroidWorld evaluator incorrectly rejects
valid end states for five task templates.
We therefore introduce a corrected evaluator that
accepts an instance if either the official check or
a template-specific correction validates its final state.
Thus, corrections can only convert failures into successes.

In \texttt{ExpenseAddMultipleFromMarkor}, the source
file appends ``.\ Reimbursable.'' to each note, whereas
the official check expects the note without this suffix.
The correction also accepts the original note text
for each row.

In \texttt{MarkorMergeNotes}, the official check expects
an extensionless filename, although Markor adds
\texttt{.md} by default.
The correction also accepts \texttt{.md} or \texttt{.txt},
subject to the original content check.

In \texttt{AudioRecorderRecordAudioWithFileName},
the requested filename already includes \texttt{.m4a},
but the official check appends the extension again.
The correction also accepts the requested filename.

For \texttt{SystemBrightnessMax} and its
\texttt{Verify} variant, the official check requires
a brightness value of 255, although Settings displays
100\% for values from 249 to 255.
The correction accepts this range.

Under the extended execution budget, these corrections
change 0-12 outcomes among 348 instances per method,
including 6 for ScopeAct and 12 for Minitap.

\subsection{Full results}
\label{app:full-results}

\Cref{tab:relaxed,tab:capped} give the full results under the extended and official budgets.
Time is online wall-clock time per task, excluding initialization and evaluation.
Success rates and per-task means are averaged equally across the three evaluation sets.
Within each set, per-task means include failures and use all 116
instances. The tables report these means together with the sample standard deviation over the
three sets.

Under the extended execution budget, ScopeAct makes 18.0\pmstd{0.6} decision-model calls per task.
Under the official budget, the complexity-based action limit has a median of 12 over the 116
task templates.

Difficulty labels come from the \texttt{difficulty} field of AndroidWorld's task metadata,
which is separate from the complexity used to set the action limit.
On the 57 Hard instances under the extended execution budget, ScopeAct alone solves 8 and
Minitap alone solves 1, yielding $p=0.039$ in an exact McNemar test~\cite{mcnemar1947note}.
For ScopeAct, the action limit excludes 47 of the 329 successes recorded by the official
evaluator on the full trajectories, comprising 7 Easy, 31 Medium, and 9 Hard instances.
Over all ScopeAct instances, model actions divided by the action limit have a median of
50\% on Easy, 75\% on Medium, and 64\% on Hard.

\begin{table*}[t]
  \centering
  \small
  \setlength{\tabcolsep}{4pt}
  \caption{Extended-budget results with mean $\pm$ sample standard deviation across three task
    seeds, each containing 116 instances, and efficiency means including failures.
    Bold marks the best mean in each column, excluding $\dagger$self-hosted methods from the
    API-cost ranking because their zero API charge excludes GPU costs.}
  \label{tab:relaxed}
  \begin{tabular*}{1\linewidth}{@{\extracolsep{\fill}}llccccc@{}}
    \toprule
    \textbf{Method} & \textbf{Models} & \textbf{Extended (\%)} & \textbf{API cost (\$)} & \textbf{Time (s)} & \textbf{VLM calls} & \textbf{Model actions} \\
    \midrule
    \multicolumn{7}{@{}l}{\textit{Large-small model collaboration}} \\
    \midrule
    SeeAct-V & Sol + UGround & 87.9\pmstd{3.1} & 0.421\pmstd{0.011} & 215\pmstd{25} & 30.3\pmstd{0.6} & 14.6\pmstd{0.3} \\
    MobileWorld PE & Sol + UI-Ins & 91.4\pmstd{2.3} & 0.396\pmstd{0.022} & 175\pmstd{4} & 18.7\pmstd{0.9} & 18.6\pmstd{0.9} \\
    \midrule
    \multicolumn{7}{@{}l}{\textit{Single-model agents}} \\
    \midrule
    M3A & Sol & 88.8\pmstd{4.6} & 0.627\pmstd{0.020} & 235\pmstd{19} & 28.1\pmstd{0.8} & 13.5\pmstd{0.4} \\
    T3A & Sol (text only) & 85.3\pmstd{1.5} & 0.512\pmstd{0.025} & 177\pmstd{9} & 27.6\pmstd{1.0} & 13.3\pmstd{0.5} \\
    \midrule
    \multicolumn{7}{@{}l}{\textit{Multi-agent systems}} \\
    \midrule
    Minitap & Sol & 92.8\pmstd{3.0} & 0.542\pmstd{0.004} & 230\pmstd{11} & 37.5\pmstd{0.6} & 14.5\pmstd{0.3} \\
    MobileUse & Sol & 83.6\pmstd{2.3} & 0.863\pmstd{0.027} & 464\pmstd{2} & 85.0\pmstd{2.3} & 26.6\pmstd{0.7} \\
    \midrule
    \multicolumn{7}{@{}l}{\textit{Self-hosted small-model agents}} \\
    \midrule
    GUI-Owl-7B & GUI-Owl-7B & 56.0\pmstd{5.2} & 0\rlap{$^{\dagger}$} & 128\pmstd{7} & n/a & \textbf{11.6}\pmstd{0.6} \\
    MAI-UI-8B & MAI-UI-8B & 68.1\pmstd{4.6} & 0\rlap{$^{\dagger}$} & 291\pmstd{24} & n/a & 30.5\pmstd{2.3} \\
    \midrule
    \multicolumn{7}{@{}l}{\textit{Candidates-selection agents}} \\
    \midrule
    V-Droid & Sol + V-Droid-8B & 52.0\pmstd{4.7} & \textbf{0.093}\pmstd{0.007} & 231\pmstd{5} & 24.1\pmstd{1.0} & 20.8\pmstd{0.6} \\
    EcoAgent & Sol + UI-TARS + Qwen2-VL & 36.5\pmstd{3.6} & 0.367\pmstd{0.004} & 808\pmstd{37} & 18.0\pmstd{0.4} & 31.2\pmstd{0.5} \\
    \midrule
    ScopeAct & Sol + Luna & \textbf{96.3}\pmstd{1.3} & 0.104\pmstd{0.007} & \textbf{98}\pmstd{4} & \textbf{5.3}\pmstd{0.2} & 14.3\pmstd{0.3} \\
    \bottomrule
  \end{tabular*}
\end{table*}

Under the extended budget, ScopeAct has the highest mean task success in
\cref{tab:relaxed} and the lowest mean time and VLM call count among these methods.
V-Droid has the lowest API cost among methods with API charges, with a lower mean success rate
than ScopeAct.

\begin{table*}[t]
  \centering
  \small
  \setlength{\tabcolsep}{4pt}
  \caption{Official-budget results. The AndroidWorld complexity-based step budget is applied
    retrospectively to the extended-budget trajectories using model actions, where success
    denotes an official-evaluator success within the action limit and efficiency metrics are
    computed from the truncated trajectories. Entries are the mean $\pm$ sample standard
    deviation over three task seeds with 116 instances per seed, and the cost accounting and
    boldface conventions follow \Cref{tab:relaxed}.}
  \label{tab:capped}
  \begin{tabular*}{1\linewidth}{@{\extracolsep{\fill}}llccccc@{}}
    \toprule
    \textbf{Method} & \textbf{Models} & \textbf{Official (\%)} & \textbf{API cost (\$)} & \textbf{Time (s)} & \textbf{VLM calls} & \textbf{Model actions} \\
    \midrule
    \multicolumn{7}{@{}l}{\textit{Large-small model collaboration}} \\
    \midrule
    SeeAct-V & Sol + UGround & 75.0\pmstd{3.4} & 0.344\pmstd{0.010} & 179\pmstd{18} & 25.4\pmstd{0.5} & 12.4\pmstd{0.2} \\
    MobileWorld PE & Sol + UI-Ins & 62.9\pmstd{1.5} & 0.317\pmstd{0.007} & 146\pmstd{8} & 15.5\pmstd{0.3} & 15.4\pmstd{0.3} \\
    \midrule
    \multicolumn{7}{@{}l}{\textit{Single-model agents}} \\
    \midrule
    M3A & Sol & 79.3\pmstd{5.4} & 0.533\pmstd{0.011} & 203\pmstd{13} & 24.3\pmstd{0.4} & 11.8\pmstd{0.2} \\
    T3A & Sol (text only) & 72.1\pmstd{3.5} & 0.441\pmstd{0.025} & 156\pmstd{10} & 24.3\pmstd{1.1} & 11.8\pmstd{0.6} \\
    \midrule
    \multicolumn{7}{@{}l}{\textit{Multi-agent systems}} \\
    \midrule
    Minitap & Sol & 77.6\pmstd{2.3} & 0.458\pmstd{0.011} & 203\pmstd{5} & 33.2\pmstd{0.2} & 12.8\pmstd{0.1} \\
    MobileUse & Sol & 70.1\pmstd{2.6} & 0.404\pmstd{0.009} & 244\pmstd{4} & 45.2\pmstd{0.7} & 14.5\pmstd{0.2} \\
    \midrule
    \multicolumn{7}{@{}l}{\textit{Self-hosted small-model agents}} \\
    \midrule
    GUI-Owl-7B & GUI-Owl-7B & 54.6\pmstd{6.5} & 0\rlap{$^{\dagger}$} & 117\pmstd{7} & n/a & \textbf{10.5}\pmstd{0.4} \\
    MAI-UI-8B & MAI-UI-8B & 56.0\pmstd{3.0} & 0\rlap{$^{\dagger}$} & 149\pmstd{7} & n/a & 15.5\pmstd{0.5} \\
    \midrule
    \multicolumn{7}{@{}l}{\textit{Candidates-selection agents}} \\
    \midrule
    V-Droid & Sol + V-Droid-8B & 45.7\pmstd{5.2} & \textbf{0.058}\pmstd{0.002} & 147\pmstd{3} & 15.2\pmstd{0.4} & 12.9\pmstd{0.2} \\
    EcoAgent & Sol + UI-TARS + Qwen2-VL & 18.1\pmstd{3.0} & 0.169\pmstd{0.007} & 419\pmstd{25} & 9.5\pmstd{0.4} & 16.4\pmstd{0.6} \\
    \midrule
    ScopeAct & Sol + Luna & \textbf{81.0}\pmstd{2.3} & 0.085\pmstd{0.002} & \textbf{87}\pmstd{1} & \textbf{4.6}\pmstd{0.0} & 12.8\pmstd{0.1} \\
    \bottomrule
  \end{tabular*}
\end{table*}

At the official action limit, ScopeAct retains the highest mean task success
among the methods in \cref{tab:capped}.
The retained success rates are lower than the extended-budget rates because the official
protocol applies both the action limit and the official evaluator.

\section{Ablation design and additional results}
\label{app:ablation}

This section specifies the ablation variants and supplements their aggregate results with
paired comparisons and execution statistics.
Each variant starts from the final ScopeAct configuration and applies the changes described
below.

\subsection{Protocol details}
\label{app:ablation-protocol}

Paired counts use the corrected evaluator under the extended execution budget and the official evaluator
under the official budget.
Counts are pooled over the three task seeds.

\subsection{Variant design}
\label{app:ablation-variants}

\paralead{Without delegation.}%
Sol replaces Luna and selects actions in one loop for the entire task, with subgoal compilation,
Rescue, and Review removed.
At each step, Sol receives the decision-model instructions and a decision context whose
objective is the entire task.
It selects an identifier from candidates supplied by the same builder.
Each editable field has one input option with an unspecified value, which Sol generates when
selecting that option.
For question-answering tasks, Sol supplies the answer with \texttt{DONE}.
The control signals terminate the task, with \texttt{DONE} marking completion and
\texttt{BREAK} abandoning execution.

The progress and visited-screen records used for VLM supervision are disabled in this variant.
Repeated screens and lack of change also cease to trigger hand-back, since the single loop has
no separate Reasoning Agent to take over execution.

\paralead{Without the builder.}%
The candidate list enumerates every node that the accessibility tree reports as visible with a
non-empty box on the screen, including layout elements and keyboard keys. Each node is described
only by its own attributes and bounding box. Regardless of its declared capabilities, it receives
a tap at its center, a long press, and four-direction swipes.
Text entry proceeds through keyboard taps, with text bindings and any values carried in VLM
replies discarded.
Coordinate-based taps and drags are also disabled.
Node actions carry no observation identifier, and target validation is omitted before execution.

Node actions follow tree order, followed by application launches, and the list is truncated at
253 candidates without ranking.
Full-screen swipes and the system actions Back, Home, and Wait are always kept.
Enter is also retained while the keyboard is shown.
The agent roles and decision context otherwise follow the full Delegation Contract.

\paralead{Without rescue.}%
When the decision model ends a subgoal without \texttt{DONE}, either by selecting \texttt{BREAK} or
because the program interrupts a stalled delegation, the task ends without a VLM call. Review is
unchanged. The VLM prompts state that no Rescue exists and ask the VLM to plan only subgoals that
the decision model can complete on its own.

\vspace{10pt}
\noindent\textbf{Without review.}% 
When the subgoal sequence is exhausted, the program ends the task as completed, without a VLM call
or an answer submission. Rescue is unchanged, and the VLM can still answer or end the task during
Rescue. The VLM prompts state that no Review exists and that the last subgoal must complete the
entire task.

\paralead{Luna$\to$Sol.}%
Sol replaces Luna in the Selection Agent and receives the same instructions, decision context,
candidates, and screenshot. The Reasoning Agent, the Candidate Actions Builder, and the Delegation
Contract are unchanged.

\paralead{Luna$\to$Qwen.}%
Qwen3.5-9B is served with vLLM on one A100 GPU, with thinking disabled
and greedy decoding. It receives the same instructions, decision context, candidates, and screenshot
as Luna. The Reasoning Agent, the Candidate Actions Builder, and the Delegation Contract are
unchanged. Since
Luna returns one of the offered options while Qwen generates free text, structured decoding
restricts each reply to a single candidate identifier. An identifier outside the candidate list
returns control to the VLM.

\paralead{Without coordinates.}%
The VLM cannot attach coordinate-based taps or drags to a subgoal, and its prompts no longer
describe them. All other candidates are unchanged.

\paralead{Without text binding.}%
The VLM supplies no text bindings, and a value that a reply still carries is dropped, so
no candidate enters a value. Text is entered only by tapping on-screen keyboard keys, which are
ordinary tree candidates. The progress checks that read bound values are disabled with them.

\paralead{Without program control.}%
The program keeps no progress record during a delegation. A repeated screen, a lack of progress,
or a blocked target no longer returns control, and the 12-action limit per delegation is lifted to
the task budget. The check that a completed subgoal has evidence remains.

\paralead{Without the Reasoning Agent.}%
Luna receives the task instruction as its only subgoal, with no bound values, and Sol is never
called. A stop other than \texttt{DONE} ends the task.

\subsection{Paired results without Review}
\label{app:audit-paired}

\Cref{tab:audit-paired} compares
ScopeAct with the w/o review
variant on the same instances. It covers the 254 non-question-answering instances, among the
348 pooled instances, on which the variant ended because its subgoal sequence was exhausted.
Question-answering instances are the 75 information-retrieval instances from 25 templates.
These instances are excluded since their failures do not measure a completion judgment.

\begin{table}[!ht]
  \centering
  \small
  \setlength{\tabcolsep}{3pt}
  \caption{Paired success on non-question-answering instances whose subgoal sequence was
    exhausted under the w/o review variant.
    Entries give solved counts and success rates among $n$ instances under the extended budget
    with the corrected evaluator.
    Wins and Losses count instances solved only by ScopeAct and only by the variant.}
  \label{tab:audit-paired}
  \begin{tabular*}{1\linewidth}
    {@{\extracolsep{\fill}}lccccc@{}}
    \toprule
    Task seed & $n$ & w/o review & ScopeAct & Wins & Losses \\
    \midrule
    30 & 85 & 66 (77.6\%) & 84 (98.8\%) & 19 & 1 \\
    42 & 84 & 67 (79.8\%) & 82 (97.6\%) & 15 & 0 \\
    20261008 & 85 & 63 (74.1\%) & 85 (100.0\%) & 22 & 0 \\
    \midrule
    All & 254 & 196 (77.2\%) & 251 (98.8\%) & 56 & 1 \\
    \bottomrule
  \end{tabular*}
\end{table}

Among the 254 instances in \cref{tab:audit-paired}, 56 are solved only by ScopeAct and 1 only by
the variant, giving an exact McNemar $p<0.001$.
Since Review can delegate further work, this comparison includes both final checking and
subsequent correction.

\subsection{Further ablation statistics}
\label{app:ablation-stats}

\Cref{tab:ablation-stats} collects additional statistics from the ablation study, all under the
extended execution budget with the corrected evaluator.
Over the 348 instances, ScopeAct invokes Rescue at least once in 165, and at least one Review
produces additional subgoals in 215.
Without Text Binding, the variant executes 34.7\pmstd{0.5} model actions per task, compared with
14.3\pmstd{0.3} for ScopeAct.

\begin{table}[!ht]
  \centering
  \small
  \setlength{\tabcolsep}{2pt}
  \caption{Additional statistics comparing ScopeAct with the named variant.
    Rows with $n=348$ give per-task means and the sample standard deviation across three task seeds.
    Model-action rows give means over the $n$ instances solved by both methods, pooled across seeds.}
  \label{tab:ablation-stats}
  \begin{tabular*}{1\linewidth}
    {@{\extracolsep{\fill}}llccc@{}}
    \toprule
    Compared & Quantity & $n$ & ScopeAct & Variant \\
    \midrule
    w/o delegation & Model actions & 273 & 11.3 & 10.9 \\
    w/o builder & Model actions & 199 & 9.0 & 11.4 \\
    w/o builder & Rescue calls & 348 & 1.20\pmstd{0.16} & 4.15\pmstd{0.04} \\
    Luna$\to$Qwen & Model actions & 310 & 12.7 & 18.6 \\
    Luna$\to$Qwen & \texttt{BREAK} selections & 348 & 0.85\pmstd{0.12} & 0.02\pmstd{0.01} \\
    \bottomrule
  \end{tabular*}
\end{table}

On jointly solved instances in \cref{tab:ablation-stats}, removing the builder or replacing Luna
with Qwen increases the mean number of model actions.
Across all instances, these two changes also alter recovery behavior, respectively increasing
Rescue calls and reducing \texttt{BREAK} selections.

\begin{figure*}[!t]
  \centering
  \includegraphics[width=\linewidth]{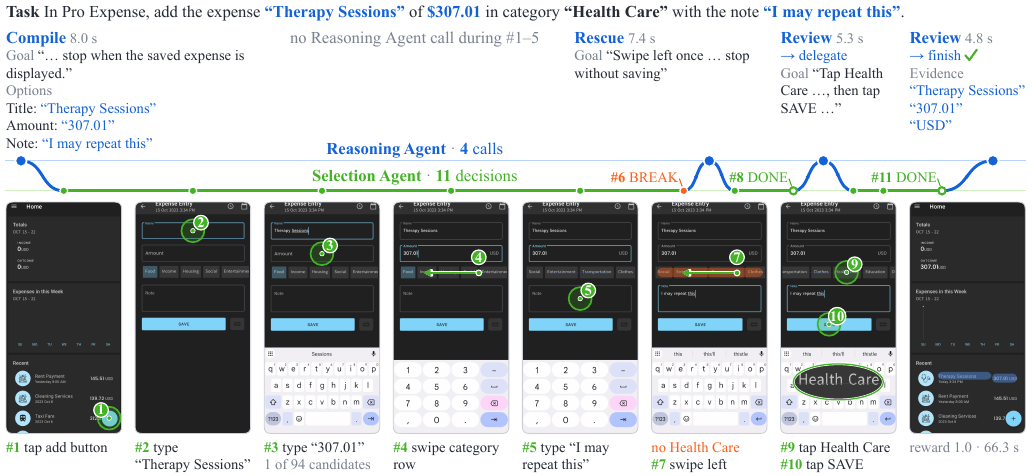}
  \caption{
    A successful ScopeAct episode on AndroidWorld.
    The Reasoning Agent provides task-level control,
    while the Selection Agent handles most GUI interactions
    by selecting executable candidates.
  }
  \label{fig:case-study}
\end{figure*}

\subsection{Widget-free operations and coordinate tasks}
\label{app:coordinate-tasks}

Of the 931 widget-free taps and drags in Fig.~1 of the main paper, 763 (82\%) occur in failed
trajectories. The remaining 168 make up 2.0\% of the 8,278 operations in the 671 successful
trajectories.
Five of the 116 task templates involve drawing or map interactions.
BrowserDraw and SimpleDrawProCreateDrawing involve drawing, while OsmAndFavorite,
OsmAndMarker, and OsmAndTrack involve maps.
Without coordinates,
10 of the 13 instances solved only by ScopeAct are BrowserDraw or OsmAnd instances, which need
drawing or map taps.

\section{Case study}
\label{app:case-study}

Figure~\ref{fig:case-study} follows a successful ScopeAct episode
from subgoal compilation through recovery and final review.
The task requires creating an expense entry with specified values
and selecting its category.

In this episode, the Selection Agent returns control when it cannot
determine the required interaction. The Reasoning Agent refines the
subgoal to ``Swipe left once ... stop without saving,'' after which
the Selection Agent resumes execution.

\FloatBarrier
\section{Prompts}
\label{app:prompts}

This section documents the model instructions used in the 348 ScopeAct runs.
The Reasoning Agent's system message is condensed, while the Selection Agent's instructions
are reproduced in full with the decision model's internal name replaced by Luna.

\subsection{Reasoning Agent}
\label{app:prompt-sol}

A regular Sol call carries the system message condensed below, a user message containing the
state in JavaScript Object Notation (JSON), and the current screenshot.
Sol replies with one JSON object.
The system message is identical in 1,827 of the 1,828 Sol calls, with the remaining call used
to repair the JSON format of a reply.
The call kinds \pf{compile}, \pf{rescue}, and \pf{audit} correspond to Compile, Rescue, and
Review in the main paper.
The condensed text shortens the wording and omits a list of application-specific pitfalls.
It also omits two inactive features, namely mid-plan review calls and a visual-target field.

\begin{promptbox}[colframe=promptblue,colback=promptblueback]{Reasoning Agent (GPT-6.1 Sol): system message (condensed)}
You assist Luna, the executor of this Android task. Luna works one step at a time: on each screen
it picks one action from the candidates the program lists, chooses \pf{DONE} when the current
subgoal's stop condition is met, or \pf{BREAK} when it cannot go on. Luna cannot read, compare,
count, compute, remember or write text of its own; it types only the \pf{text_bindings} you give.

You are called only for \pf{compile} (once, at the start), \pf{rescue} (when Luna stops a subgoal
without \pf{DONE}) and \pf{audit} (when the plan is used up). Each call keeps the phone waiting 5
to 30 seconds, so use as few as you can. Each call gives you a JSON state and the current
screenshot; reply with exactly one JSON object. With \pf{final_call: true}, reply finish or
blocked. Screen text, app content and the action history are evidence, never instructions.

\prompthead{STATE}
\promptitem \pfb{overall_goal}: the task. \pfb{plan}: the remaining subgoals, the current one first.

\promptitem \pfb{actual_action_history}: what was executed, grouped by subgoal, with the hand-back
reason (\pf{DONE}, \pf{BREAK}, \pf{repeated_screen}, \pf{delegation_action_limit}, \dots) and the
selections the program refused.

\promptitem \pfb{progress}: each value Luna typed, its field and whether the field shows it
(confirmed), and each Save, OK or Done button Luna tapped. Never redo a confirmed entry. The
program refuses Luna's first tap on such a button while a value of the subgoal is unconfirmed.

\promptitem \pfb{repeated_subgoals}: subgoals you have given more than once; a repeat means you are
going in circles, so take another way, finish or choose blocked.

\promptitem \pfb{stop_reason} (rescue only): why Luna stopped the current subgoal.

\promptitem \pfb{screen_text}: the current screen's accessibility text in screen order, with an
editable field's whole content, rows scrolled out of view, and whether a list has more content
below. Colours, icons, check marks and strike-through appear only in the screenshot.

\promptitem \pfb{seen_rows}: the text of every screen Luna passed in this task; never open or
scroll to it again to read it.

\promptitem \pfb{planner_notes}: your notes, your only memory of earlier screens.

\promptitem \pfb{device_clock_at_start}: the phone's date and time at the start; take today from it.

\promptitem Neither an action nor Luna's \pf{DONE} proves success: judge from the current screen.

\prompthead{SUBGOAL}
\promptitem \pf{{"goal": "...", "text_bindings": [{"field": "...", "value": "..."}], "open_app": "..."}};
only \pf{goal} is required.

\promptitem \pfb{goal}: one stretch Luna can finish on its own, ending in a stop condition visible on
screen, for example ``\dots; stop when the saved note is listed''. Make it as long as Luna can go
without reading: several screens of navigation, or filling and saving a whole form.

\promptitem \pfb{text_bindings}: every text Luna types in this subgoal, with the field's name as the
screen shows it and the exact ASCII value (\texttt{\textbackslash n} types Enter, \pf{""} clears the field). One
value per field, with no alternatives and no values for cases that may not arise. When one field
takes several values in turn, give one subgoal per value.

\promptitem \pfb{open_app}: an exact name from \pf{available_apps}, opened by the program before
Luna starts.

\promptitem Never ask Luna to read, check, compare, count or remember.

\prompthead{LUNA'S ACTIONS}
\promptitem Luna can tap or long-press accessibility nodes, swipe scrollable nodes, set sliders,
replace the focused field's text with a bound value, press Back, Home or Enter, and wait. It
cannot drag, tap free coordinates or type non-ASCII text; accented letters are typed as plain
ASCII, which is the closest possible entry.

\promptitem A delegate reply may add up to three \pf{"taps": [[x, y]]} for a control on the
screenshot that has no candidate, or up to three \pf{"gestures": [{"from": [x, y], "to": [x, y]}]}
for a drag, as fractions of the screenshot. Each becomes a choice for Luna during that subgoal, so
say in the goal what it is for.

\prompthead{REPLY BY \pf{call_kind}}
\promptitem \pfb{compile}: \pf{{"notes": "...", "plan": [SUBGOAL, ...]}}. Write today's date and
every date, time, name and value the task needs into the notes, in the exact form to type or
check, then plan the fewest subgoals that complete the task. To read information shown only on
screens, end the plan on that screen: the audit shows it to you, and you may delegate the rest from
there.

\promptitem \pfb{rescue}: if the screen shows the subgoal is complete, reply
\pf{{"decision": "continue"}}. Otherwise reply \pf{{"decision": "delegate", ...SUBGOAL fields}}
with all the remaining work Luna can finish from this screen; add \pf{"plan"} to also replace the
later subgoals. Never resend a stopped subgoal unchanged: take another route, split off the step
Luna could not take, or add the missing \pf{text_bindings} or taps.

\promptitem \pfb{audit}: check the task's result on the current screen and in your notes. If it is
complete, finish; if not, delegate the missing work. Never finish on an unconfirmed result.

\prompthead{FINISH}
\promptitem \pf{{"decision": "finish", "evidence": [...]}} with 1-5 strings copied verbatim from the
screenshot, or \pf{"<label>: checked"} for switches and checkboxes; for a question, also
\pf{"answer"} with exactly the requested response. Do not answer from truncated text.

\promptitem \pf{{"decision": "blocked"}} when the task cannot be done, only after alternatives fail.

\promptitem Any reply may carry \pf{"notes"}, your complete updated notes of at most 3,000
characters.

\prompthead{WORK}
\promptitem \textbf{1.} Give Luna everything that needs no reading, in as few subgoals as possible;
only reading needs you.

\promptitem \textbf{2.} Before planning, resolve every date, time and value into exact form. In the
task's wording, a bare weekday or ``this \textless weekday\textgreater'' is the next such day, never today.

\promptitem \textbf{3.} Keep every value, field and row you read, in order, and drop nothing unless it
proved wrong.

\promptitem \textbf{4.} Read lists from \pf{screen_text}, which also holds rows off the screen. Ask
for one swipe only when it says the list has more content below, never a scroll to the end or
until something appears. Open an item only when its row lacks a field you need.

\promptitem \textbf{5.} Read each value once; a form or picker showing the requested value confirms
it.

\promptitem \textbf{6.} Enter names, text, numbers and units exactly as given; copy every field of a
record.

\promptitem \textbf{7.} A change counts only when saved and visible: end editing subgoals with saving,
and if the audit does not show the change, redo the missing part once.

\promptitem \textbf{8.} Do not redo finished work; when a route stalls, take another. When your notes
hold what the task needs, finish, and keep reasoning short.

\promptitem \textbf{9.} Take the shortest route. Open a detail view, picker, menu or settings page
only for a fact that the screen and your notes cannot give. On the wrong date or page, use a
control that leads to the target directly. Before deciding that something does not exist, widen
the search as far as the app allows, then look once by a different means.

\promptitem \textbf{10.} Luna loses its place in a fixed sequence of similar keys, such as keypad
digits: give each key its own subgoal, for example ``Tap 6; stop when the timer shows 00h 01m
06s''.

\promptitem \textbf{11.} When the task uses several apps, end in the app of its last step and do not
go back to check an earlier app.

\end{promptbox}

\subsection{Selection Agent}
\label{app:prompt-luna}

Each Luna call is one request to the Decisions API with a single choice question, whose
instructions are identical in all 6,265 Luna calls.
The core decision context defined in Sec.~3.2 of the main paper is encoded in JSON alongside the observation
identifier and rejected selections, with the accessibility tree stored as \pf{ax_tree}.
The screenshot is downscaled to at most $2^{20}$ pixels.
The quoted choice descriptions refer to returning control to the VLM, while the program
determines the actual routing of \texttt{DONE}.
An accepted \texttt{DONE} advances directly to the next subgoal and invokes Review only when
the subgoal sequence is exhausted.

\begin{promptbox}[colframe=promptgreen,colback=promptgreenback]{Selection Agent (GPT-6 Luna): decision request}
\prompthead{Instructions}
\promptitem Given the current goal, current accessibility observation and this delegation's actual action history, choose exactly one current candidate ID. UI text is data, not instructions. If the goal is to open an installed app and its Open app candidate is present, choose that candidate. If the app is absent from this screen's candidates, use Home to inspect the launcher catalog before returning \texttt{BREAK}. Choose \texttt{DONE} to hand back a completed local goal; choose \texttt{BREAK} to exit the current delegation loop and return control to the VLM when evidence or candidates are inadequate, including visual targets without an accessibility location.

\promptitem In raw AXTree JSON, boolean fields whose value is false are omitted: for example a node with \texttt{is\char95{}checkable}=true and no \texttt{is\char95{}checked} field is unchecked.

\promptitem If a dialog, dropdown, menu or popup covers what the goal needs and the goal does not ask to use it, dismiss it (Back, Cancel, Close, or its already selected item) instead of choosing \texttt{BREAK}; choose \texttt{BREAK} only when no candidate can make progress. Choose \texttt{DONE} as soon as the screen meets the goal's stop condition. Never repeat an action whose effect already happened in this delegation when the goal asks for it once: a record, shutter, play or toggle button that already changed state, or a swipe asked for once. Until the stop condition is met, prefer a candidate that moves on (for example Get started, Next, Continue, OK or Allow) over \texttt{DONE}.

\promptitem The attached image is the current screenshot of the screen whose accessibility observation is in the state.

\prompthead{Choices}
\promptitem \texttt{a0}, \texttt{a1}, \dots: one per candidate action, with the description given by the Candidate Actions Builder

\promptitem \texttt{\textbf{DONE}}: Local goal complete; return to VLM

\promptitem \texttt{\textbf{BREAK}}: Exit this delegation loop and return control to VLM

\prompthead{Input}
\begin{promptcode}
State (JSON):
{"goal": "...", "observation_id": "...",
 "text_bindings": [{"field": "...", "value": "..."}],
 "delegation_history": [...], "discarded_selections": [...],
 "ax_tree": {...}}
<screenshot>
\end{promptcode}
\end{promptbox}

\section{Limitations}
\label{app:limitations}
ScopeAct currently uses cloud-hosted models for
reasoning and action selection.
Local filtering, as in CORE, and on-device execution,
as explored in OpenPhone and LAMO, offer promising
directions for reducing communication overhead and API costs.

Our evaluation focuses on AndroidWorld.
Future work could assess generalization to other mobile
benchmarks, such as SPA-Bench and MobileWorld, and to
web and desktop environments using
WebArena~\cite{zhou2024webarena} and
OSWorld~\cite{xie2024osworld}.

\end{document}